\documentclass[pdflatex,sn-nature]{sn-jnl}

\usepackage{graphicx}
\usepackage{booktabs}
\usepackage{amsmath,amssymb}
\usepackage{xcolor}
\usepackage{tabularx}
\usepackage{array}
\usepackage{url}
\usepackage{float}

\begin{document}

\title[Aperiodic reliance in physiological deep learning]{A spectral audit framework reveals task-dependent aperiodic reliance across EEG and ECG deep learning}

\author*[1]{\fnm{Jasmeet Singh} \sur{Bindra}}\email{D24059@students.iitmandi.ac.in}
\author[2]{\fnm{Siddharth} \sur{Panwar}}\email{siddharth.panwar@iitmandi.ac.in}

\affil*[1]{\orgdiv{Indian Knowledge Systems and Mental Health Applications (IKSMHA) Center}, \orgname{Indian Institute of Technology Mandi}, \orgaddress{\state{Himachal Pradesh}, \country{India}}}
\affil[2]{\orgdiv{School of Computing and Electrical Engineering}, \orgname{Indian Institute of Technology Mandi}, \orgaddress{\state{Himachal Pradesh}, \country{India}}}

\abstract{
Deep learning on physiological time series is interpreted through domain-specific features---oscillatory rhythms in EEG, morphological complexes in ECG---yet these signals sit atop a broadband aperiodic 1/f-like envelope that covaries with arousal, age, and pathology. We introduce a spectral audit framework combining aperiodic/periodic decomposition, phase-preserving Fourier interventions, sham controls, and simulation validation. Aperiodic reliance was task-dependent and architecture-general: across six neural architectures, flattening drops exceeded 0.42 balanced-accuracy points for sleep-wake classification, reached 0.07-0.13 for clinical abnormality detection, and remained minimal for motor imagery. Six of seven EEG foundation models showed FDR-significant aperiodic reliance on clinical EEG; age/sex and recording-era controls reduced but did not eliminate the effect. Applying the audit to PTB-XL ECG revealed neural drops of 0.32--0.36 persisting after demographic matching, confirming this confound class extends beyond EEG. Aperiodic controls should become standard for interpretable physiological time-series deep learning.
}


\maketitle

Electroencephalography (EEG) deep learning models are increasingly used for sleep staging, clinical abnormality detection and brain--computer interface decoding \cite{supratak2017deepsleepnet,chambon2018deep,phan2019seqsleepnet,perslev2019utime,lawhern2018eegnet,schirrmeister2017deep,gemein2020machine}. Their performance is often interpreted using the language of oscillatory physiology: delta activity for sleep depth, alpha and mu rhythms for sensorimotor state, beta activity for cognitive or motor processes \cite{buzsaki2004neuronal,buzsaki2012origin,achermann1997low,pfurtscheller2001motor}. This interpretation is intuitive and historically grounded, but it assumes that the spectral features used by a model correspond primarily to oscillatory peaks or band-limited power.

EEG power spectra also contain a broadband aperiodic component that follows an approximately 1/f-like profile \cite{donoghue2020parameterizing,wen2016separating,voytek2015age}. This component is not a nuisance constant. Its offset and exponent vary with vigilance, age, pathology and recording conditions \cite{lendner2020electrophysiological,voytek2015age,brake2024neurophysiological}. Classical band-power analyses can therefore confound oscillatory changes with shifts in the broadband spectral envelope \cite{donoghue2020parameterizing,gao2017inferring}. The same issue is potentially more consequential for deep learning: modern models can exploit any predictive regularity in the input, including spectral slope, amplitude distribution and waveform morphology, while their outputs are often interpreted as if they had discovered oscillatory biomarkers. The same risk extends to any physiological modality whose power spectrum contains a predictive broadband envelope, including electrocardiography.

High-performing EEG and ECG deep learning models can rely on broadband aperiodic spectral structure rather than the oscillatory or morphological biomarkers invoked to interpret them---and this reliance is task-dependent. We developed a spectral audit framework that decomposes spectra into aperiodic and periodic components, applies matched phase-preserving Fourier interventions to raw signals, and quantifies aperiodic sufficiency and necessity with subject-level uncertainty. We evaluated the framework across three EEG domains \cite{kemp2000sleep,kemp2013sleepedf,obeid2016tuh,schalk2004bci2000,goldberger2000physionet} using six standard architectures \cite{lawhern2018eegnet,schirrmeister2017deep}, seven EEG foundation models \cite{yang2023biot,jiang2024labram,yue2024eegpt,wang2025cbramod,elouahidi2025reve,wang2025eegmamba,kostas2021bendr} and PTB-XL ECG as a cross-modality proof of principle \cite{wagner2020ptbxl,wang2017timeseries,fawaz2020inceptiontime}. Aperiodic reliance was dominant in sleep arousal, present and partly age-confounded in clinical EEG, minimal in lateralized motor imagery, and strongly present in ECG abnormality detection.

\section{Results}

\subsection{A spectral audit framework for EEG deep learning}

\begin{figure}[t]
\centering
\includegraphics[width=\textwidth]{"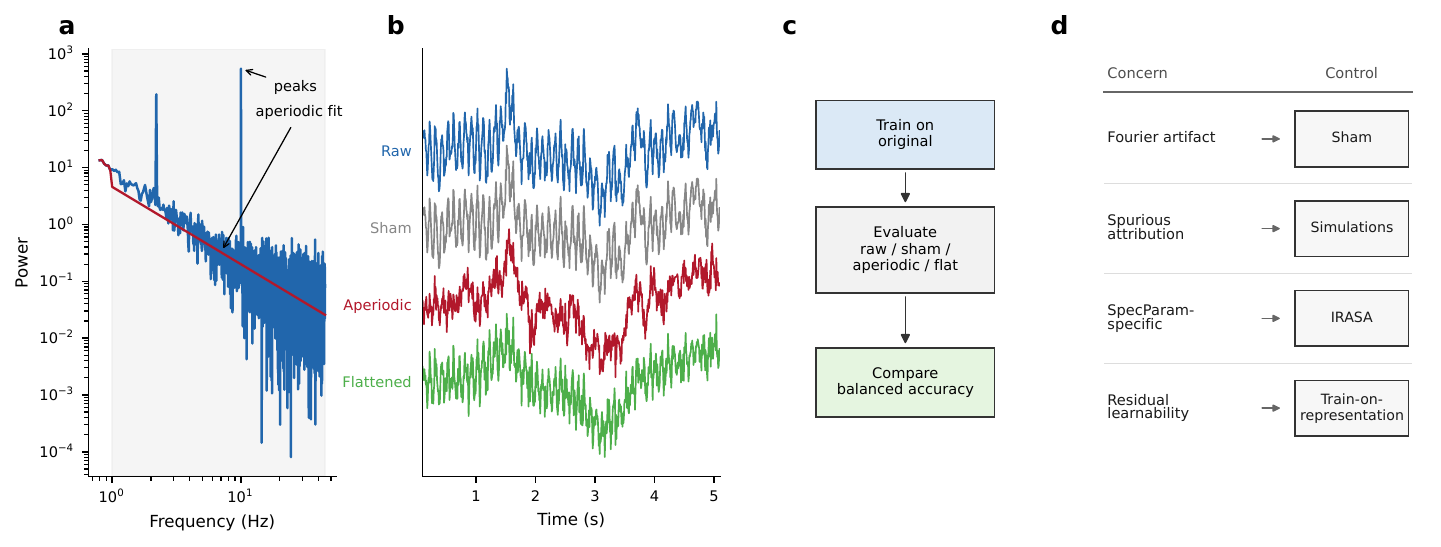"}
\caption{\textbf{Spectral audit framework.} \textbf{a}, A representative EEG power spectrum is decomposed into full, aperiodic-shaped and flattened residual inputs. \textbf{b}, Raw EEG interventions preserve Fourier phase while modifying amplitude spectra to construct sham, aperiodic-shaped and flattened time-domain signals. \textbf{c}, Models are trained on original data and evaluated under matched intervention conditions. \textbf{d}, The validation stack combines sham reconstruction, simulation validation, IRASA agreement and train-on-representation controls.}
\label{fig:framework}
\end{figure}

The audit begins by estimating each epoch or trial power spectrum and decomposing it into a broadband aperiodic envelope and residual periodic structure (Fig.~\ref{fig:framework}a) \cite{donoghue2020parameterizing,wen2016separating}. For PSD-based models, we compare three representations: the full log-PSD, the aperiodic-shaped spectrum and the flattened residual spectrum obtained after removing the aperiodic envelope. These tests ask whether aperiodic structure is sufficient for classification and whether performance survives when the aperiodic component is suppressed.

For raw EEG models, we use phase-preserving Fourier interventions (Fig.~\ref{fig:framework}b). The raw signal is Fourier transformed, its phase is retained, and its amplitude spectrum is replaced by one of three matched alternatives: a sham reconstruction that preserves the original spectrum, an aperiodic-shaped spectrum that removes narrowband peaks, or a flattened spectrum that suppresses the broadband envelope. This design keeps the temporal phase structure fixed while manipulating the spectral information available to the classifier. Models are trained on the original condition and evaluated on all intervention conditions (Fig.~\ref{fig:framework}c). The full validation stack includes sham controls, synthetic simulations, decomposition agreement and train-on-representation controls (Fig.~\ref{fig:framework}d) \cite{gerster2022separating}.

Synthetic validation confirmed that the audit behaves as intended. When class labels were generated only by aperiodic differences, full-spectrum and aperiodic-spectrum classifiers performed near ceiling while flattened spectra collapsed to chance. When labels were generated by periodic peaks, flattened residual spectra retained information. Under mixed and shortcut-shift simulations, the intervention matrix tracked the injected ground truth (Fig.~\ref{fig:controls}b; Supplementary Note 2).

\subsection{Aperiodic reliance is task- and domain-dependent}

\begin{figure}[t]
\centering
\includegraphics[width=\textwidth]{"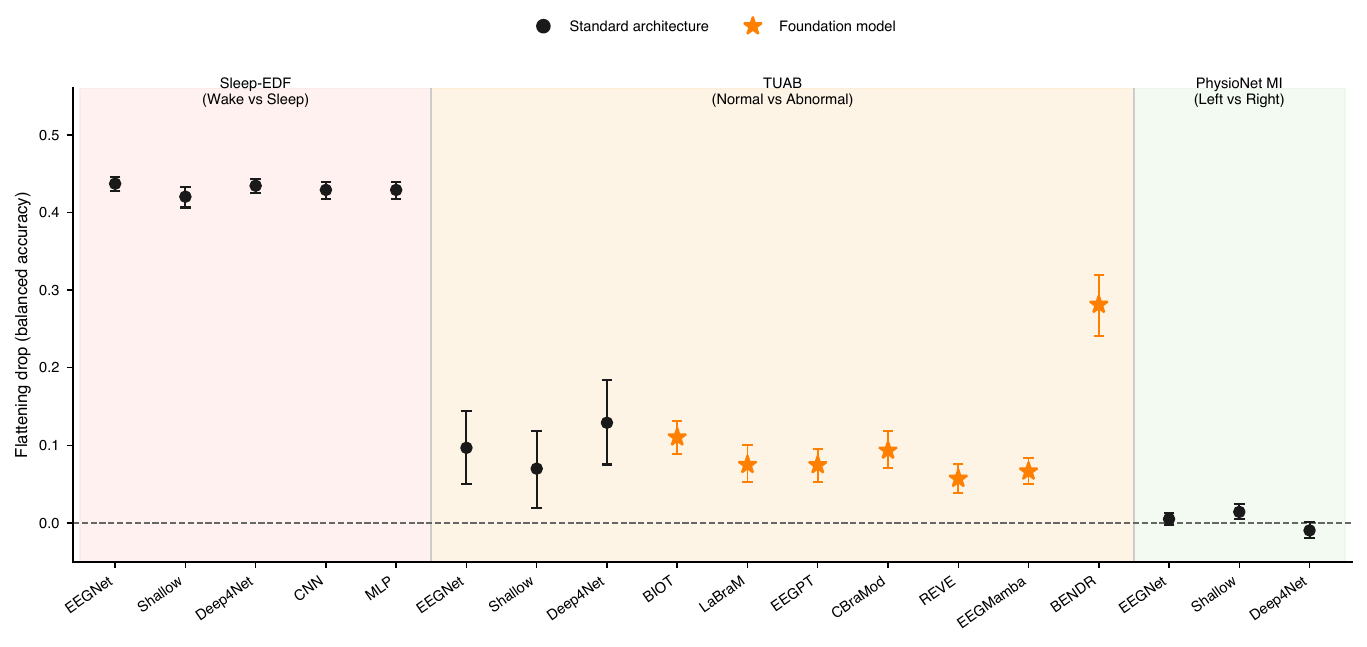"}
\caption{\textbf{Aperiodic reliance varies across EEG domains.} A flattening drop in balanced accuracy is shown for the primary task in each dataset (wake versus sleep, normal versus abnormal and left versus right imagery, respectively). Sleep-EDF wake-versus-sleep classification shows large drops across architectures. Full-TUAB abnormality detection shows intermediate drops in standard neural models and EEG foundation models. PhysioNet motor imagery shows near-zero drops. Error bars denote 95\% subject-level bootstrap confidence intervals.}
\label{fig:crossdomain}
\end{figure}

Across domains, flattening the aperiodic envelope produced a clear gradient of model vulnerability (Fig.~\ref{fig:crossdomain}; Table~\ref{tab:summary}). Sleep-EDF wake-versus-sleep classification showed the largest dependence. EEGNet dropped by 0.437 balanced-accuracy points after flattening, with similarly large drops for ShallowFBCSPNet, Deep4Net, a CNN and the MLP baseline. This behavior was expected if the model uses the strong arousal-related broadband structure separating wakefulness and sleep \cite{lendner2020electrophysiological,bodizs2021composite}.

Full-TUAB abnormality detection showed an intermediate pattern \cite{obeid2016tuh,gemein2020machine}. Across 253 evaluation subjects, EEGNet, ShallowFBCSPNet and Deep4Net lost performance after flattening, but the drops were smaller than in Sleep-EDF: 0.097, 0.070 and 0.129 balanced-accuracy points, respectively. Foundation models were not exempt. In a seven-model, three-seed TUAB audit, six models showed FDR-significant flattening drops, ranging from 0.057 for REVE to 0.110 for BIOT, while BENDR was a cautionary negative-transfer case because the sham intervention itself collapsed performance to chance \cite{yang2023biot,jiang2024labram,yue2024eegpt,wang2025cbramod,elouahidi2025reve,wang2025eegmamba,kostas2021bendr}. The expected specificity checks were non-significant for Sleep-EDF N2-versus-N3 and for PhysioNet MI EEGNet and Deep4Net, whereas PhysioNet MI ShallowFBCSPNet showed a small FDR-significant drop (0.014; $p_{\mathrm{FDR}}=0.0037$). We therefore describe motor imagery as showing minimal, rather than zero, aperiodic reliance \cite{schalk2004bci2000,pfurtscheller2001motor,wolpaw2002brain,blankertz2008optimizing}. Across the formal test family, 25 of 31 primary flattening tests survived Benjamini--Hochberg correction, showing that aperiodic reliance is widespread in sleep and clinical EEG audits while remaining minimal in the expected specificity cases.

\begin{table*}[t]
\centering
\caption{\textbf{Cross-domain summary of primary intervention results.} Balanced accuracy is reported with 95\% subject-level bootstrap confidence intervals. The primary task is wake versus sleep for Sleep-EDF, normal versus abnormal for full TUAB v3.0.1 and left versus right fist imagery for PhysioNet MI. TUAB neural and foundation-model rows use three-seed hierarchical seed/subject bootstrap aggregation.}
\label{tab:summary}
\tiny
\setlength{\tabcolsep}{2pt}
\renewcommand{\arraystretch}{1.12}
\begin{tabularx}{\textwidth}{>{\raggedright\arraybackslash}p{0.09\textwidth}>{\raggedright\arraybackslash}p{0.10\textwidth}>{\raggedright\arraybackslash}p{0.10\textwidth}>{\centering\arraybackslash}p{0.12\textwidth}>{\centering\arraybackslash}p{0.12\textwidth}>{\centering\arraybackslash}p{0.12\textwidth}>{\centering\arraybackslash}p{0.12\textwidth}>{\centering\arraybackslash}p{0.13\textwidth}}
\toprule
Domain & Task & Model & Raw BA & Sham BA & Aperiodic BA & Flattened BA & Flattening Drop \\
\midrule
Sleep-EDF & Wake vs Sleep & EEGNet & 0.939 [0.929, 0.948] & 0.939 [0.929, 0.948] & 0.879 [0.860, 0.897] & 0.502 [0.500, 0.505] & \textbf{0.437 [0.427, 0.446]} \\
Sleep-EDF & Wake vs Sleep & \shortstack[l]{Shallow\\FBCSPNet} & 0.931 [0.920, 0.941] & 0.931 [0.920, 0.941] & 0.858 [0.833, 0.882] & 0.511 [0.505, 0.518] & \textbf{0.420 [0.406, 0.432]} \\
Sleep-EDF & Wake vs Sleep & Deep4Net & 0.941 [0.931, 0.948] & 0.941 [0.931, 0.948] & 0.890 [0.872, 0.907] & 0.506 [0.503, 0.510] & \textbf{0.434 [0.425, 0.443]} \\
Sleep-EDF & Wake vs Sleep & CNN & 0.940 [0.932, 0.948] & 0.940 [0.932, 0.948] & 0.871 [0.853, 0.889] & 0.511 [0.505, 0.520] & \textbf{0.429 [0.417, 0.439]} \\
Sleep-EDF & Wake vs Sleep & MLP & 0.929 [0.918, 0.939] & NA & 0.783 [0.757, 0.808] & 0.500 [0.500, 0.500] & \textbf{0.429 [0.418, 0.439]} \\
\midrule
TUAB & Normal vs Abnormal & EEGNet & 0.804 [0.765, 0.843] & 0.804 [0.765, 0.843] & 0.578 [0.491, 0.658] & 0.707 [0.655, 0.758] & \textbf{0.097 [0.050, 0.144]} \\
TUAB & Normal vs Abnormal & \shortstack[l]{Shallow\\FBCSPNet} & 0.796 [0.755, 0.836] & 0.796 [0.755, 0.836] & 0.575 [0.511, 0.639] & 0.727 [0.676, 0.774] & \textbf{0.070 [0.020, 0.119]} \\
TUAB & Normal vs Abnormal & Deep4Net & 0.816 [0.777, 0.854] & 0.816 [0.777, 0.854] & 0.574 [0.512, 0.635] & 0.687 [0.633, 0.739] & \textbf{0.129 [0.075, 0.184]} \\
TUAB & Normal vs Abnormal & BIOT & 0.802 [0.781, 0.823] & 0.800 [0.779, 0.821] & 0.678 [0.651, 0.704] & 0.692 [0.666, 0.719] & \textbf{0.110 [0.088, 0.132]} \\
TUAB & Normal vs Abnormal & LaBraM & 0.786 [0.763, 0.809] & 0.786 [0.763, 0.809] & 0.696 [0.674, 0.722] & 0.711 [0.683, 0.739] & \textbf{0.075 [0.052, 0.100]} \\
TUAB & Normal vs Abnormal & EEGPT & 0.801 [0.779, 0.824] & 0.801 [0.778, 0.823] & 0.685 [0.660, 0.712] & 0.727 [0.702, 0.752] & \textbf{0.075 [0.053, 0.096]} \\
TUAB & Normal vs Abnormal & CBraMod & 0.770 [0.746, 0.795] & 0.770 [0.746, 0.795] & 0.640 [0.612, 0.672] & 0.677 [0.644, 0.709] & \textbf{0.093 [0.071, 0.119]} \\
TUAB & Normal vs Abnormal & REVE & 0.780 [0.753, 0.806] & 0.780 [0.753, 0.806] & 0.675 [0.643, 0.702] & 0.723 [0.696, 0.749] & \textbf{0.057 [0.038, 0.076]} \\
TUAB & Normal vs Abnormal & EEGMamba & 0.781 [0.756, 0.807] & 0.781 [0.756, 0.806] & 0.693 [0.668, 0.716] & 0.715 [0.690, 0.739] & \textbf{0.066 [0.050, 0.083]} \\
TUAB & Normal vs Abnormal & BENDR$^\dagger$ & 0.781 [0.742, 0.818] & 0.500 [0.487, 0.513] & 0.500 [0.487, 0.513] & 0.500 [0.487, 0.513] & \textbf{0.281 [0.241, 0.320]} \\
\midrule
PhysioNet MI & Left vs Right imagery & EEGNet & 0.744 [0.720, 0.767] & 0.744 [0.720, 0.767] & 0.737 [0.714, 0.760] & 0.739 [0.717, 0.761] & 0.005 [-0.002, 0.013] \\
PhysioNet MI & Left vs Right imagery & \shortstack[l]{Shallow\\FBCSPNet} & 0.655 [0.634, 0.678] & 0.655 [0.634, 0.678] & 0.617 [0.597, 0.639] & 0.641 [0.618, 0.665] & \textbf{0.014 [0.005, 0.024]} \\
PhysioNet MI & Left vs Right imagery & Deep4Net & 0.682 [0.661, 0.703] & 0.682 [0.661, 0.703] & 0.684 [0.663, 0.705] & 0.692 [0.669, 0.715] & -0.010 [-0.020, 0.001] \\
\bottomrule
\end{tabularx}
\vspace{2pt}
\raggedright{\scriptsize $^\dagger$BENDR is reported as a negative-transfer/intervention-fragility case because its sham condition collapsed to chance; it is not interpreted as clean aperiodic-specific reliance.}
\end{table*}

The sham intervention was essentially neutral across clean audits, with BENDR as the sole exception (Fig.~\ref{fig:controls}a). Thus, outside this flagged fragility case, the flattening effects are not explained by the Fourier reconstruction step alone. They reflect the information removed or retained by the spectral intervention.

\subsection{N2-versus-N3 classification is robust to aperiodic flattening}

\begin{figure}[t]
\centering
\includegraphics[width=\textwidth]{"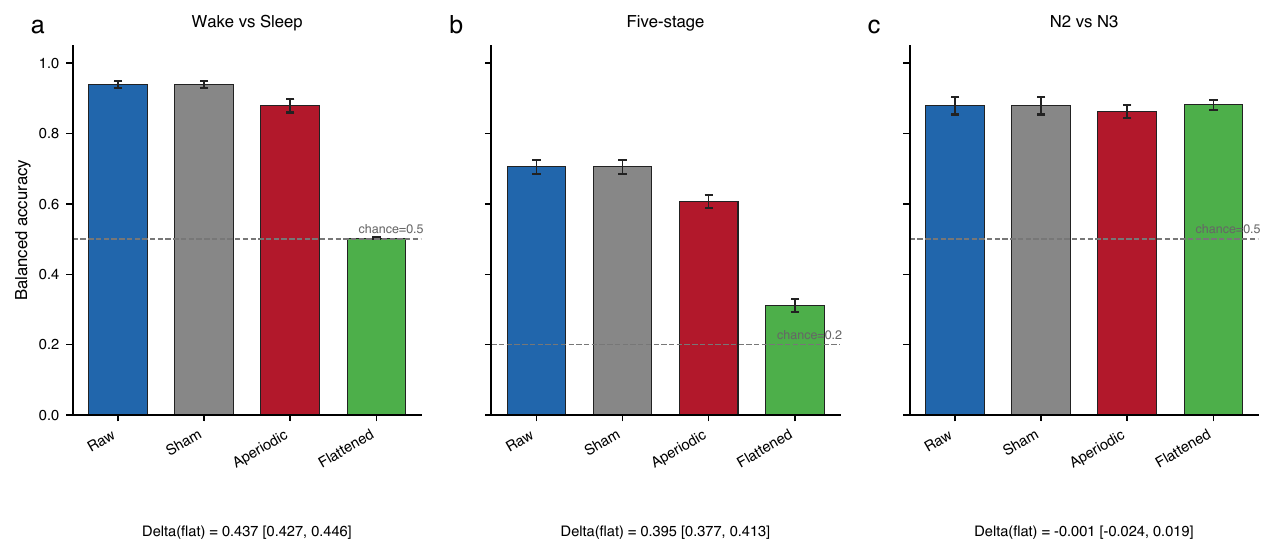"}
\caption{\textbf{Task-specific dissociation within Sleep-EDF.} EEGNet intervention results are shown for wake-versus-sleep, five-stage staging and N2-versus-N3 classification. Flattening strongly disrupts wake-versus-sleep and five-stage performance but leaves N2-versus-N3 performance intact. This dissociation indicates that the intervention is diagnostic rather than generically destructive.}
\label{fig:sleep}
\end{figure}

Within Sleep-EDF, the audit separated tasks that superficially belong to the same domain. Wake-versus-sleep and five-stage sleep staging showed large EEGNet flattening drops, but N2-versus-N3 classification did not (Fig.~\ref{fig:sleep}). EEGNet achieved similar balanced accuracy on raw and flattened inputs for N2-versus-N3, with a near-zero drop. The same qualitative pattern appeared across architectures in the extended Sleep-EDF results, and all four tested N2-versus-N3 flattening drops were non-significant after FDR correction (all $p_{\mathrm{FDR}}>0.4$).

This exception is important for interpretation. If flattening merely damaged time-domain signals or erased all useful EEG information, all sleep tasks would collapse. Instead, N2-versus-N3 remains decodable after flattening, consistent with models using slow-wave morphology or residual temporal structure rather than a global spectral slope \cite{achermann1997low,massimini2004sleep,bodizs2021composite}. The audit therefore distinguishes between aperiodic reliance and non-aperiodic sleep physiology.

\subsection{Clinical EEG classification relies on age-confounded aperiodic structure}

TUAB is the strongest test of the framework because the target is clinically important and demographic confounding is plausible \cite{obeid2016tuh}. In the full TUAB v3.0.1 corpus, abnormal subjects were older than normal subjects, while aperiodic structure remained highly predictive of abnormality. This creates an immediate interpretive ambiguity: a model may detect clinical abnormality, age-related spectral slope or both \cite{voytek2015age,brake2024neurophysiological,demuru2020fingerprinting}. Disentangling these sources is critical because aperiodic structure may be simultaneously a demographic shortcut and a genuine biomarker of neural dysfunction.

\begin{figure}[t]
\centering
\includegraphics[width=\textwidth]{"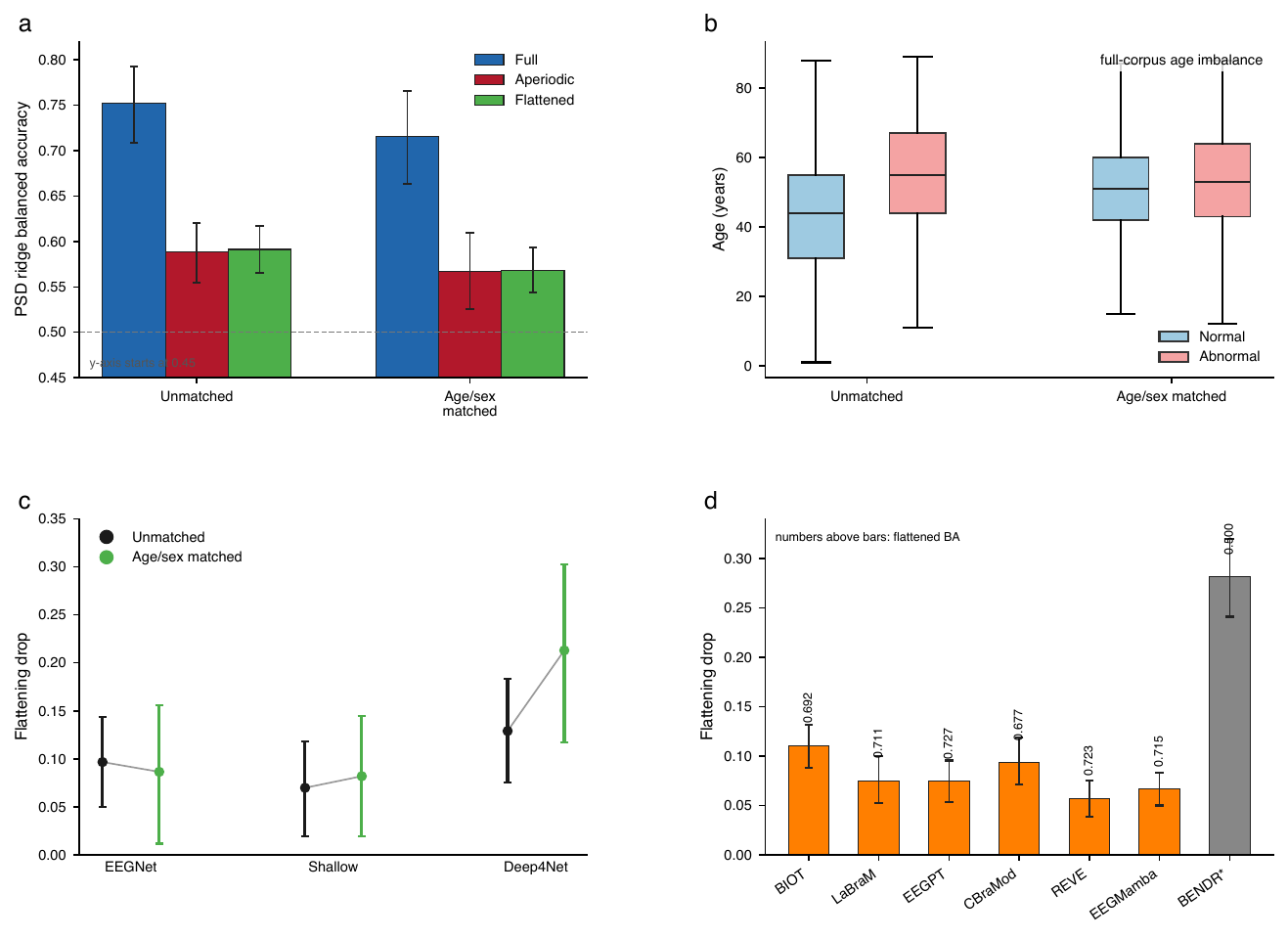"}
\caption{\textbf{TUAB clinical benchmark audit.} \textbf{a}, Full-TUAB PSD ridge results before and after age/sex matching. \textbf{b}, Age distributions show the abnormal--normal age imbalance in the full corpus and its reduction after matching. The y-axis in performance panels is truncated to emphasize differences near the operating range. \textbf{c}, Full-TUAB neural-model flattening drops before and after age/sex matching. \textbf{d}, Foundation-model flattening drops for seven full-TUAB audits. BIOT, LaBraM, EEGPT, CBraMod, REVE and EEGMamba show clean positive drops; BENDR is flagged because sham performance collapsed to chance and is therefore not interpretable as clean aperiodic reliance.}
\label{fig:tuab}
\end{figure}

The PSD ridge audit showed that full-spectrum TUAB performance reached 0.752 balanced accuracy and fell by 0.160 after flattening. We then paired abnormal and normal subjects within official TUAB split boundaries using same-sex matches and a five-year age caliper, yielding 921 matched pairs: 834 training pairs and 87 evaluation pairs, covering 174 evaluation subjects. In this matched control, the PSD ridge model retained above-chance full-spectrum performance (0.715) and showed equal aperiodic and flattened drops of 0.148. Neural flattening drops also persisted after matching: 0.087 for EEGNet, 0.082 for ShallowFBCSPNet and 0.213 for Deep4Net (Fig.~\ref{fig:tuab}c). This argues against a simple explanation in which TUAB performance is only an age shortcut. Instead, aperiodic structure appears to contain both demographic and pathology-related information.

The clinical implication is two-sided. Aperiodic structure may be a real biomarker of abnormal EEG, but it is also entangled with age. Reported TUAB performance therefore should not be interpreted as purely pathology-specific unless demographic controls and aperiodic controls are included.

A temporal acquisition-proxy audit tested whether the TUAB effect could instead be explained by recording-era drift. Using EDF recording dates, we split the usable corpus into early (2009--2011) and late (2012--2013) eras. PSD ridge flattening drops persisted within early and late recordings (0.118 and 0.204), and also generalized across eras when training early and evaluating late (0.146) or training late and evaluating early (0.164). Year-tercile sensitivity checks were consistent (Extended Data Fig.~10), indicating that acquisition period modulates effect size but does not explain away the TUAB aperiodic result.

\subsection{Foundation models encode exploitable aperiodic structure}

Large-scale pretraining did not remove aperiodic reliance. Across the full TUAB evaluation set, BIOT showed the largest clean foundation-model flattening drop (0.110) and LaBraM showed a smaller but significant drop (0.075), consistent with LaBraM retaining more non-aperiodic information \cite{yang2023biot,jiang2024labram}. EEGPT also dropped by 0.075, with a larger aperiodic-shaped drop than flattened drop, suggesting a mixture of aperiodic and residual oscillatory or spatiotemporal evidence \cite{yue2024eegpt}. CBraMod showed clear aperiodic reliance (drop 0.093), while REVE and EEGMamba showed smaller but still significant drops of 0.057 and 0.066 \cite{wang2025cbramod,elouahidi2025reve,wang2025eegmamba}. Thus, all six clean foundation-model flattening drops were formally significant after FDR correction (all $p_{\mathrm{FDR}}<0.0001$).

BENDR was different \cite{kostas2021bendr}. Its baseline balanced accuracy was 0.781, but sham, aperiodic and flattened interventions all collapsed to 0.500. BENDR was the only FDR-significant sham-control result ($p_{\mathrm{FDR}}<0.0001$), confirming it as an intervention-fragility case rather than evidence of clean aperiodic-specific reliance. A plausible mechanism is that the BENDR convolutional encoder is unusually sensitive to exact amplitude statistics or waveform details altered by Fourier round-tripping, even when the amplitude spectrum is nominally preserved in the sham condition. This failure mode is informative: public foundation checkpoints can look viable under raw evaluation while being unstable under controlled spectral perturbation.

This result matters for benchmark interpretation. Foundation models are often evaluated by accuracy improvements over task-specific architectures, but accuracy alone does not reveal whether the representation uses disease physiology, demographic shortcuts, broadband recording statistics or oscillatory patterns. The same audit can be applied after a foundation model's preprocessing and before its forward pass, making aperiodic intervention a practical diagnostic for future pretrained EEG models. Current EEG foundation-model benchmarks report only raw accuracy; adding an aperiodic intervention column would reveal whether apparent gains reflect learned neurophysiology or broadband statistical shortcuts.

\subsection{The spectral audit generalizes to ECG time series}

The same confound class should appear whenever a time-series domain contains predictive 1/f-like spectral structure. We therefore applied the audit to PTB-XL normal-versus-abnormal electrocardiography as a non-EEG physiological proof of principle \cite{wagner2020ptbxl,goldberger2000physionet}. The prepared cohort contained 21,375 ten-second, 12-lead ECG records from 18,610 patients (one of the largest publicly available diagnostic ECG resources), with fold 10 held out for testing. A PSD ridge baseline reached 0.707 balanced accuracy and fell to 0.607 after flattening, a drop of 0.100 [0.079, 0.122].

Established ECG/time-series neural architectures showed larger effects. ResNet1D-Wang, Inception1D and XResNet1D101 achieved raw balanced accuracies of 0.860, 0.866 and 0.860, respectively; sham reconstructions were neutral, but flattened inputs reduced performance to 0.512, 0.508 and 0.538, corresponding to drops of 0.347, 0.358 and 0.322 (Extended Data Fig.~9) \cite{wang2017timeseries,fawaz2020inceptiontime}. PTB-XL also showed a TUAB-like demographic imbalance: abnormal records were older than normal records (mean age 64.95 versus 52.04 years). Same-sex, five-year-caliper matching reduced the PSD ridge flattening drop to 0.032 [0.009, 0.056], indicating that much of the PSD-level effect was demographic. By contrast, matched neural drops remained large: 0.312, 0.311 and 0.299. Thus, deep ECG models exploited broadband or morphology-linked spectral structure beyond the age/sex-correlated PSD slope. The dissociation between PSD-level demographic sensitivity and neural-level persistence mirrors the TUAB pattern, suggesting deep models access morphological or waveform-level broadband features beyond what linear spectral slope captures.

The ECG extension is not a definitive ECG biomarker study. SpecParam fit quality was substantially lower for ECG than EEG (median channel-level $R^2=0.273$), reflecting QRS harmonics and heart-rate periodicity. It nevertheless demonstrates that the confound class identified in EEG is not EEG-specific: any physiological time series with a predictive broadband envelope is susceptible to the same failure mode.

\subsection{Decomposition agreement and methodological controls}

\begin{figure}[t]
\centering
\includegraphics[width=\textwidth]{"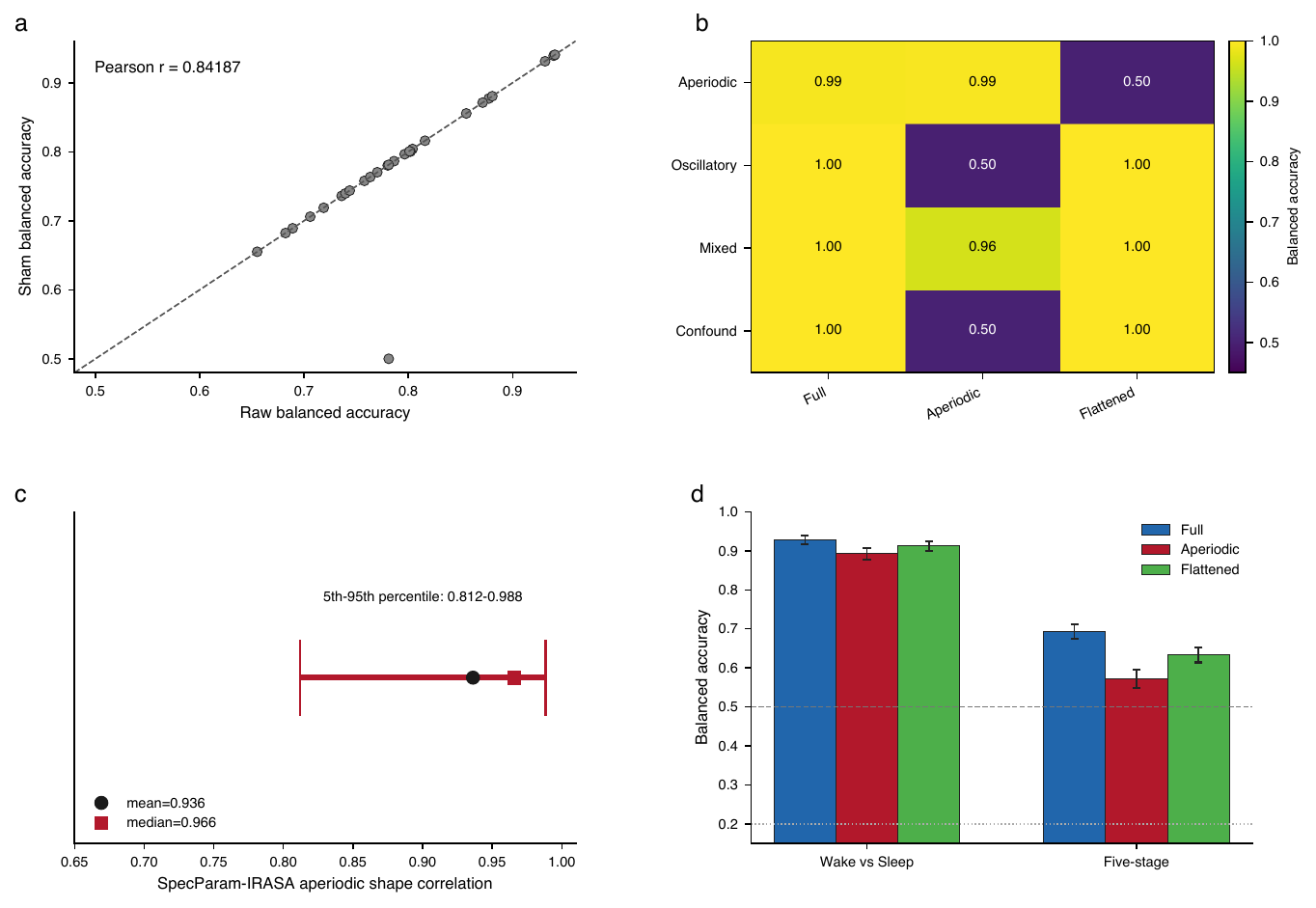"}
\caption{\textbf{Validation and controls.} \textbf{a}, Sham reconstruction preserves model performance across domain--task--architecture combinations. \textbf{b}, Simulation validation recovers known ground-truth dependence on aperiodic and periodic signal generators. \textbf{c}, IRASA and SpecParam aperiodic shapes agree strongly on Sleep-EDF. \textbf{d}, Train-on-representation controls show that flattened spectra can retain usable information when a model is trained on that representation, separating lack of learned reliance from lack of residual signal.}
\label{fig:controls}
\end{figure}

Several controls address whether the observed effects are artifacts of the decomposition or intervention procedure (Fig.~\ref{fig:controls}). SpecParam and IRASA produced strongly concordant aperiodic shapes on Sleep-EDF, with median shape correlation 0.966 \cite{donoghue2020parameterizing,wen2016separating,gerster2022separating}. This agreement between two methodologically independent decomposition approaches indicates that the audit's conclusions are not artifacts of a particular spectral fitting algorithm. The IRASA-derived ridge intervention reproduced the main qualitative result: aperiodic-shaped spectra retained substantial sleep-stage information and flattened spectra removed the dominant wake-versus-sleep and five-stage cues.

Train-on-flattened controls further clarify the PSD result. Flattening a full-PSD model's test input can collapse performance because the model learned a dominant aperiodic axis. But models trained directly on flattened PSDs recover residual information in some tasks, showing that flattening does not universally destroy all signal. Together, the sham, simulation, IRASA, and train-on-representation controls support the conclusion that the audit measures model reliance on aperiodic structure rather than generic preprocessing damage.

\section{Discussion}

These results show that aperiodic spectral structure is a measurable, task-dependent contributor to EEG deep learning performance and a broader time-series confound. In sleep arousal classification, removing the aperiodic envelope reduces neural model performance by more than 0.4 balanced-accuracy points, consistent with prior links between spectral slope and arousal state \cite{lendner2020electrophysiological,bodizs2021composite}. In full TUAB, it contributes to conventional neural networks and six clean foundation-model audits, while motor imagery shows minimal reliance. In PTB-XL ECG, the same audit reveals large neural flattening drops that persist after age/sex matching.

The central implication is interpretive. A model can be accurate while relying on information that is not the oscillatory mechanism invoked in the discussion of a paper. This does not make such models invalid. It means that claims about oscillatory biomarkers require explicit aperiodic controls \cite{donoghue2020parameterizing,gao2017inferring}. Aperiodic-only and flattened inputs, sham raw interventions and subject-level uncertainty are low-cost additions compared with the cost of training modern EEG models, and they substantially improve scientific interpretability. Crucially, the audit requires no retraining---only evaluation-time spectral manipulation---making its computational cost negligible relative to the models being evaluated.

The TUAB findings are especially important. The abnormal group is older, and aperiodic exponent is age-sensitive \cite{voytek2015age,demuru2020fingerprinting}. Full-corpus age/sex matching reduced the concern but did not eliminate aperiodic reliance, suggesting a mixed signal: part demographic covariation and part clinically relevant broadband physiology \cite{brake2024neurophysiological}. A temporal acquisition-proxy audit further showed that recording-era drift is not the sole explanation: flattening drops persisted within early and late recording bins and transferred across them. TUAB benchmark scores may therefore partly reflect age-correlated spectral slope, but aperiodic structure may also encode abnormality-related neurophysiology. This duality---simultaneous confound and biomarker---represents a fundamentally different interpretive challenge than simple shortcut learning. Unlike a pure confound that should be removed, aperiodic structure in clinical EEG may require disentanglement rather than elimination; the audit provides the measurement tool, and the clinical interpretation must follow.

The motor-imagery result is the complementary sanity check. If the audit declared every EEG task aperiodic-dependent, it would be too blunt to guide interpretation. Instead, PhysioNet MI remained robust after flattening across multiple neural architectures. This supports the view that the method identifies where broadband structure matters and where oscillatory explanations remain plausible.

The foundation-model results carry distinct implications. Six of seven clean audits showed FDR-significant aperiodic reliance, meaning that the current generation of pretrained EEG encoders has not solved the broadband shortcut problem through scale. This matters because foundation models are increasingly positioned as general-purpose EEG representations whose downstream clinical claims inherit whatever confound structure the encoder has absorbed. The BENDR sham collapse is a separate but complementary warning: public checkpoints can appear functional on standard benchmarks while proving brittle under controlled spectral perturbation. Foundation-model leaderboards that report only raw accuracy obscure these distinctions; adding aperiodic intervention scores as a standard evaluation axis would separate genuine representational advances from broadband exploitation.

We recommend that EEG machine-learning publications adopt four methodological controls: reporting of aperiodic decomposition fit quality for all datasets, intervention-based sufficiency and necessity tests comparing full, aperiodic-only and flattened inputs, sham controls for any raw-domain intervention, and explicit labeling of whether performance claims rest on oscillatory or broadband features. Published EEG-DL studies should re-examine interpretive claims when a model is described as using alpha, beta, delta or other band-limited biomarkers without ruling out broadband envelope dependence.

At the ecosystem level, foundation-model leaderboards should include aperiodic intervention scores next to raw accuracy, so that benchmark gains can be separated from robustness of the learned representation. Clinical translation pipelines should run the audit before deployment, especially when diagnosis, age, medication or recording state can covary with spectral slope. Dataset creators should release demographic and acquisition metadata whenever ethically permissible, because those variables make age/sex matching and site-aware controls possible.

The ECG extension elevates this work from an EEG-specific methodological contribution to a general time-series principle. PTB-XL differs from EEG in sensor modality, physiology and waveform structure, yet deep models still lost 0.32--0.36 balanced-accuracy points after flattening, and this effect survived age/sex matching--evidence that the audit targets a general spectral failure mode in which a broadband envelope covaries with labels regardless of domain. Similar risks plausibly arise in EMG, heart-rate variability, speech, financial and seismological time series, but those domains remain to be tested directly. The audit framework is domain-agnostic by design---it requires only that the signal have a decomposable power spectrum---making it immediately applicable to any 1/f-bearing time series.

There are limitations. PhysioNet MI uses short cue-locked trials, making spectral fitting more challenging than in 20--30 s epochs, although fit quality was explicitly reported. ECG SpecParam fit quality was substantially lower than EEG because QRS complexes introduce strong harmonic structure; the PTB-XL analysis should therefore be read as a proof of principle, not a definitive ECG biomarker study. BENDR also shows that not every pretrained EEG checkpoint transfers cleanly into a unified intervention pipeline: its sham collapse makes it a cautionary result rather than a clean aperiodic-reliance estimate. Finally, the intervention framework identifies reliance, not the causal biological origin of the aperiodic signal, and it diagnoses the problem without prescribing a training-time solution. Developing aperiodic-aware training objectives---for example, data augmentation via spectral flattening or adversarial regularization against broadband features---is an important direction that the current framework enables but does not address. Future work should extend the audit to additional clinical EEG domains, longitudinal disease progression, other physiological signals and domain-specific sequence models such as U-Sleep \cite{perslev2021usleep}.

The aim of the audit is not to demote deep learning, but to make its successes scientifically legible \cite{brookshire2024leakage}.

\section{Methods}

\subsection{Datasets}

Sleep analyses used the Sleep-EDF Expanded Sleep Cassette cohort with 78 subjects \cite{kemp2000sleep,kemp2013sleepedf,goldberger2000physionet}. We evaluated three tasks: wake versus sleep, five-stage sleep staging and N2 versus N3 discrimination. Epochs followed the standard 30 s sleep-staging structure. Unless otherwise stated, metrics were aggregated by subject before bootstrap confidence intervals were computed.

Clinical abnormality detection used the full TUAB v3.0.1 corpus within the official TUAB train/evaluation split \cite{obeid2016tuh,gemein2020machine}. The local full-corpus manifest contained 2,993 EDF files from approximately 2,329 unique subject identifiers; the raw-neural intervention cache contained 204,122 epochs, with 185,705 training epochs and 18,417 evaluation epochs. The official evaluation set comprised 253 subjects. The primary task was normal versus abnormal EEG classification. Header-derived demographic metadata were extracted when available and used for the age/sex-matched control.

Motor imagery analyses used the PhysioNet EEG Motor Movement/Imagery dataset with 109 subjects and 64 channels \cite{schalk2004bci2000,goldberger2000physionet}. The primary task was imagined left versus right fist movement. Because motor imagery is conventionally analyzed in short cue-locked windows rather than 30 s epochs, PSD estimates used multitaper spectra and fit quality was reported separately \cite{pfurtscheller2001motor,thomson1982spectrum}.

To test whether the audit generalizes beyond EEG, ECG analyses used PTB-XL v1.0.3 at 100\,Hz, a 12-lead ECG resource, distributed through PhysioNet \cite{wagner2020ptbxl,goldberger2000physionet}. Records with diagnostic class NORM only were labeled normal, records with any non-NORM diagnostic class were labeled abnormal and ambiguous records were excluded. The final cohort contained 21,375 ten-second records from 18,610 patients, including 9,097 normal and 12,278 abnormal records. Fold 10 was used as the held-out test set and contained 2,157 records.

\subsection{Preprocessing}

Sleep-EDF recordings were segmented into 30 s epochs using the accompanying hypnogram annotations. Wake epochs were trimmed to retain wakefulness within 30 min of sleep, matching common Sleep-EDF practice. We used the two EEG derivations Fpz--Cz and Pz--Oz, converted EDF volt units to microvolts, applied a zero-phase fourth-order Butterworth bandpass filter from 0.5--45 Hz, resampled to 100 Hz and removed the per-epoch channel mean. Epochs extending outside the PSG recording were discarded; no additional amplitude-based artifact rejection was applied.

TUAB v3.0.1 recordings were processed within the official train/evaluation split. For the standard PSD and raw-neural TUAB analyses, recordings were segmented into non-overlapping 20 s windows with 20 s stride. We used 21 referential channels: FP1, FP2, F3, F4, C3, C4, P3, P4, O1, O2, F7, F8, T3, T4, T5, T6, A1, A2, FZ, CZ and PZ. Signals were converted to microvolts, filtered from 1--45 Hz with a zero-phase fourth-order Butterworth filter, resampled to 100 Hz and mean-centred per epoch and channel. Windows with incomplete samples or missing required channels were excluded; no amplitude-threshold rejection was otherwise applied. TUAB PSDs were estimated from these 20 s windows using multitaper spectra over 1--45 Hz with 2 Hz bandwidth.

PhysioNet motor-imagery analyses used the EEG Motor Movement/Imagery imagined left-versus-right fist task. Trials were extracted from 0.5--4.0 s after each T1/T2 cue, excluding rest annotations. All 64 EEG channels were used, converted from volts to microvolts, filtered from 1--45 Hz with a zero-phase fourth-order Butterworth filter, resampled to 160 Hz and mean-centred per trial and channel. Trials extending beyond the EDF recording boundary were discarded; no additional amplitude-based rejection was applied.

PTB-XL ECG records were analyzed as ten-second, 12-lead signals at 100 Hz. Signals were filtered from 0.5--40 Hz and used directly for neural-model interventions. PSD features were estimated with Welch spectra over 1--45 Hz.

\subsection{Spectral decomposition}

Each log-power spectrum was modeled as a sum of aperiodic and periodic components \cite{donoghue2020parameterizing}:
\begin{equation}
\log_{10} S(f) \;=\; L(f) \;+\; \sum_{k} G_k(f),
\label{eq:specparam}
\end{equation}
where $L(f) = b - \chi\,\log_{10} f$ is the aperiodic component parameterized by offset $b$ and exponent $\chi$ (fixed mode, without knee), and each $G_k$ is a Gaussian peak. The aperiodic-shaped spectrum isolates the broadband envelope, $\hat{A}(f) = 10^{L(f)}$, and the flattened residual spectrum
\begin{equation}
S_{\mathrm{flat}}(f) \;=\; \frac{S(f)}{\hat{A}(f)}
\label{eq:flatten}
\end{equation}
retains periodic peak structure while suppressing the broadband slope. This division preserves relative peak amplitudes while removing their aperiodic pedestal. Fixed mode was used because knee frequencies were not reliably identifiable across all datasets and frequency ranges; sensitivity to this choice was evaluated in Extended Data Fig.~7. Sleep-EDF and TUAB used the 1--45\,Hz range; PhysioNet MI used 2--45\,Hz because of the short trial duration and coarser low-frequency resolution. SpecParam fit quality was summarized by dataset in Supplementary Table 2.

IRASA was used as an independent decomposition check on Sleep-EDF \cite{wen2016separating}. The IRASA-derived aperiodic shapes were compared with SpecParam-derived shapes using per-epoch shape correlations, and a downstream ridge intervention was repeated using IRASA-derived full, aperiodic and flattened spectra \cite{gerster2022separating}.

\subsection{PSD intervention protocol}

For PSD-based models, classifiers were trained and evaluated across a train-by-test representation matrix. The three input representations were the full log-PSD $\log_{10} S(f)$, the aperiodic-shaped spectrum $\log_{10} \hat{A}(f)$ and the flattened residual $\log_{10} S_{\mathrm{flat}}(f)$ (Eq.~\ref{eq:flatten}). PSDs were estimated using Welch or multitaper methods as appropriate for the epoch duration \cite{welch1967use,thomson1982spectrum}. A model trained on $S(f)$ and tested on $S_{\mathrm{flat}}(f)$ estimates the \emph{necessity} of the aperiodic envelope. A model trained and tested on $\hat{A}(f)$ estimates aperiodic \emph{sufficiency}. A model trained and tested on $S_{\mathrm{flat}}(f)$ estimates residual periodic information that remains after aperiodic removal.

\subsection{Raw-signal Fourier interventions}

For raw neural models, each time-domain epoch $x(t)$ was decomposed via discrete Fourier transform into amplitude and phase, $X(f)=|X(f)|e^{j\varphi(f)}$. The phase spectrum $\varphi(f)$ was preserved across all intervention conditions. Let $\mathcal{B}$ denote the intervention band. For EEG, the fitted log-power aperiodic envelope $L(f)$ was converted to a centred amplitude shape,
\begin{equation}
a(f)=10^{\frac{1}{2}\{L(f)-\langle L\rangle_{\mathcal{B}}\}},
\label{eq:amp_shape}
\end{equation}
where $\langle\cdot\rangle_{\mathcal{B}}$ denotes averaging over frequencies in $\mathcal{B}$. The sham, aperiodic-shaped and flattened Fourier coefficients were then
\begin{align}
\tilde{X}_{\mathrm{sham}}(f) &= |X(f)|e^{j\varphi(f)}, \label{eq:sham}\\[3pt]
\tilde{X}_{\mathrm{aper}}(f) &= g_X\,a(f)e^{j\varphi(f)}, \qquad
g_X=\exp\langle\log |X(f)|\rangle_{\mathcal{B}}, \label{eq:aper}\\[3pt]
\tilde{X}_{\mathrm{flat}}(f) &= \frac{|X(f)|}{a(f)}e^{j\varphi(f)}, \label{eq:flat}
\end{align}
for $f\in\mathcal{B}$, with coefficients outside $\mathcal{B}$ left unchanged. For PTB-XL ECG, the same operation was applied in the log-amplitude domain using a linear 1/f fit over the intervention band. Reconstructed signals were inverse-transformed, mean-centred and scale-matched to the original epoch:
\begin{equation}
\hat{x}_{\mathrm{matched}}(t)
=
\{\hat{x}(t)-\bar{\hat{x}}\}
\frac{\sigma_{\mathrm{orig}}}{\sigma_{\mathrm{interv}}},
\label{eq:rms}
\end{equation}
where $\sigma$ is the per-epoch, per-channel scale estimate after centring. This prevents global amplitude scale differences between conditions from driving classifier decisions.

\subsection{Model architectures}

The model suite included ridge regression for PSD baselines, a deep MLP for PSD inputs, a raw CNN, EEGNet, ShallowFBCSPNet and Deep4Net \cite{pedregosa2011scikit,lawhern2018eegnet,schirrmeister2017deep}. EEGNet, ShallowFBCSPNet and Deep4Net were implemented through Braindecode-compatible architectures to match commonly used EEG deep-learning baselines \cite{schirrmeister2017deep,gemein2020machine}. PTB-XL neural audits used three established ECG/time-series architectures: ResNet1D-Wang, Inception1D and XResNet1D101 \cite{wang2017timeseries,fawaz2020inceptiontime}. ECG neural models were trained for 35 epochs with batch size 192 and three seeds. The main cross-domain table uses the primary task for each dataset; complete numerical results are provided in Supplementary Table 1.

\subsection{Foundation-model audits}

Seven foundation models were audited on full-TUAB normal-versus-abnormal classification using the same intervention logic as the raw neural models. Each model was initialized from its public pretrained checkpoint, fine-tuned with a binary classification head on the TUAB training split, and evaluated on the official TUAB evaluation split under raw, sham, aperiodic-shaped, and flattened conditions. Interventions were applied after each model's required preprocessing and immediately before the forward pass, so that the audit tested the representation actually presented to the pretrained encoder. All foundation-model summaries used three seeds and hierarchical seed/evaluation-subject bootstrap aggregation with 10,000 resamples. Model-specific preprocessing and checkpoint details are listed in Supplementary Table 6.

\subsection{TUAB age/sex matching}

TUAB demographic metadata were extracted from EDF header fields. Age/sex matching was performed within the official TUAB split boundary using same-sex abnormal--normal pairs and a five-year age caliper. The full-corpus match produced 921 matched pairs: 834 training pairs and 87 evaluation pairs. The matched evaluation set therefore contained 174 subjects. The matched subset was used as a control, not as a replacement for the primary TUAB result.

\subsection{PTB-XL age/sex matching}

To test whether ECG aperiodic effects were demographic, PTB-XL records were matched using the same principle as TUAB. Records with the de-identified age sentinel value 300 were excluded (284 records). Within train, validation and test split groups, abnormal and normal records were paired by same sex and a five-year age caliper. This produced 5,524 training pairs, 692 validation pairs and 698 test pairs. The matched neural bootstrap used matched pair as the resampling unit.

\subsection{TUAB temporal acquisition-proxy audit}

To test whether TUAB aperiodic reliance reflected recording-era drift, EDF recording dates were extracted from header metadata and used as a non-signal acquisition proxy. Most usable recordings fell between 2009 and 2013; sentinel or outlier years 1899, 2000 and 2007 were excluded. The primary split compared early recordings (2009--2011) with late recordings (2012--2013), and a sensitivity analysis used year terciles (2009--2010, 2011--2012 and 2013). The same PSD ridge features as the main TUAB audit were evaluated within-era and across-era, with 10,000 subject-level bootstrap resamples.

\subsection{Statistical analysis}

The primary metric was balanced accuracy. Macro-F1 and accuracy were also computed for supporting analyses. Confidence intervals were computed using subject-level bootstrap aggregation, preserving the subject as the unit of uncertainty. For multi-seed neural and foundation-model experiments, hierarchical bootstrap aggregation proceeded as follows: for each of $B=10{,}000$ resamples, seed indices were sampled with replacement from $\{1,\dots,S\}$, and within each sampled seed, evaluation-subject indices were sampled with replacement from $\{1,\dots,N\}$; the metric was then computed on the resampled data and averaged across sampled seeds, yielding a bootstrap distribution that captures both seed-level and subject-level variability. Flattening drop was defined as $\Delta_{\mathrm{flat}} = \mathrm{BA}_{\mathrm{raw}} - \mathrm{BA}_{\mathrm{flat}}$, and retention as $R = \mathrm{BA}_{\mathrm{interv}}\,/\,\mathrm{BA}_{\mathrm{raw}}$.

Formal hypothesis tests were computed from the same bootstrap distributions. For each primary flattening-drop comparison, we tested $H_0:\Delta_{\mathrm{flat}}\leq 0$ against $H_1:\Delta_{\mathrm{flat}}>0$, with one-sided bootstrap $p$ values defined as the fraction of 10,000 resamples with non-positive drop. Sham controls were tested separately against $H_0:\Delta_{\mathrm{sham}}=0$ using two-sided bootstrap $p$ values. Benjamini--Hochberg correction at $q=0.05$ was applied separately to the 31 primary flattening tests and the 28 sham-control tests.

\subsection{Simulation validation}

Synthetic spectra were generated under four families to test known ground-truth dependence. Spectra were sampled over 177 frequency bins spanning 1--45\,Hz, with 40 subjects, 120 epochs per subject and two channels per family. Let $y^\star\in\{-1,+1\}$ denote the class sign. In the \emph{pure aperiodic} family, classes differed in both aperiodic offset and exponent, with expected class values $b_A=0.32$, $b_B=0.68$, $\chi_A=0.78$ and $\chi_B=1.22$, plus subject-level jitter $\epsilon_b\sim\mathcal{N}(0,0.12)$ and $\epsilon_\chi\sim\mathcal{N}(0,0.08)$; no periodic component was added. In the \emph{pure periodic} family, classes shared aperiodic parameters but differed in a signed oscillatory template, $0.45\,y^\star\{G(f;10.0,1.2)-0.65G(f;13.5,1.5)\}$, where $G(f;\mu,\sigma)$ is a Gaussian in frequency. The \emph{mixed} family combined smaller aperiodic class effects ($\Delta b=\pm0.14$, $\Delta\chi=\pm0.16$) with a periodic template weight of 0.30. The \emph{shortcut-confound} family added a train-only aperiodic shortcut ($\Delta b=\pm0.22$, $\Delta\chi=\pm0.24$) while retaining a stable periodic template weight of 0.22 at train and test time. A ridge classifier was trained and tested across the full PSD intervention matrix (full, aperiodic-shaped, flattened representations). In all four families, the intervention matrix correctly recovered the injected ground truth (Fig.~\ref{fig:controls}b; Supplementary Note 2).

\clearpage
\section*{Extended Data}

\begin{figure}[H]
\centering
\includegraphics[width=\textwidth]{"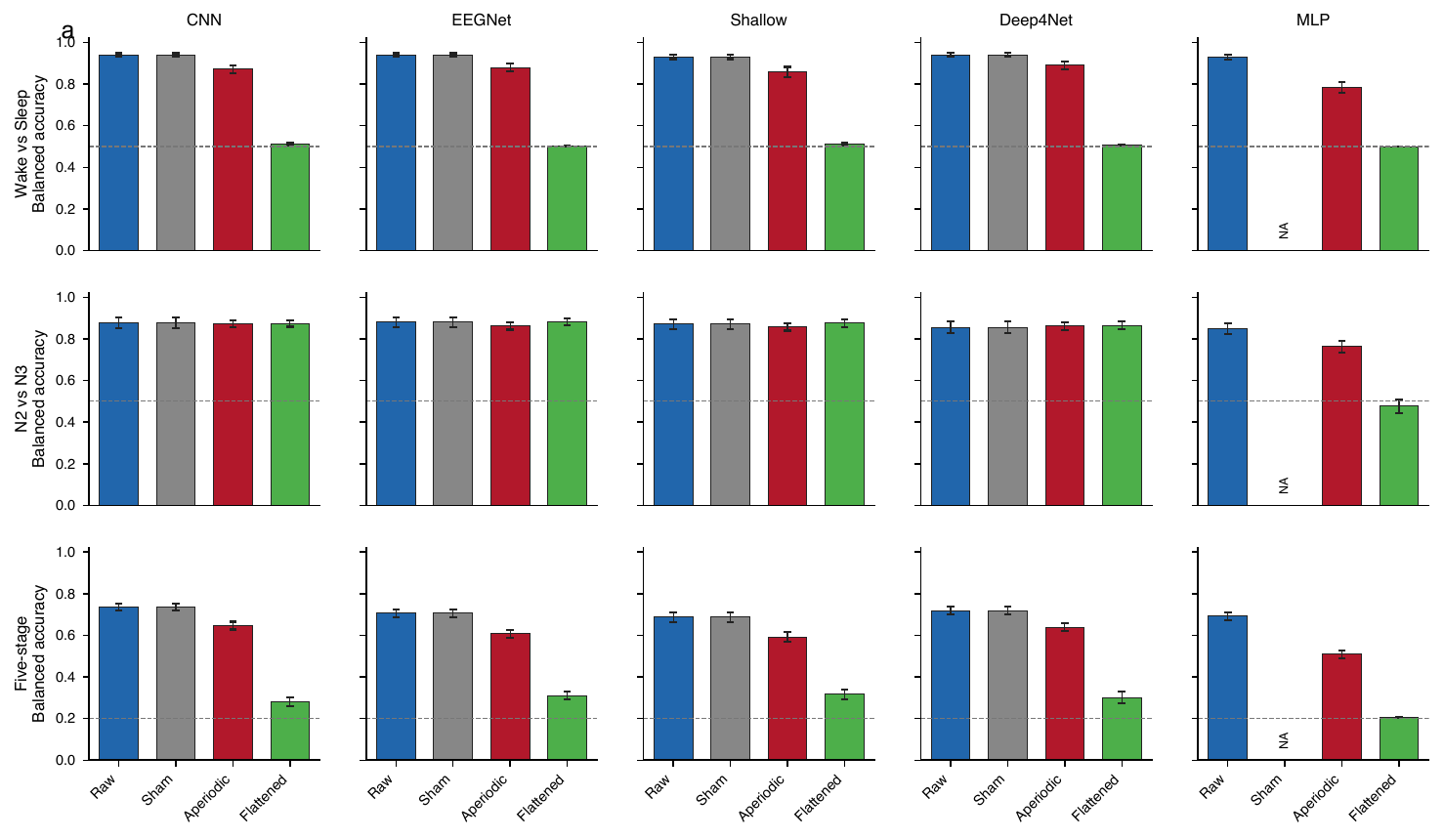"}
\caption{\textbf{Extended Data Fig.~1 $|$ Full Sleep-EDF intervention results.} Complete Sleep-EDF task-by-architecture intervention results for wake-versus-sleep, five-stage staging and N2-versus-N3 classification.}
\end{figure}
\clearpage

\begin{figure}[H]
\centering
\includegraphics[width=\textwidth]{"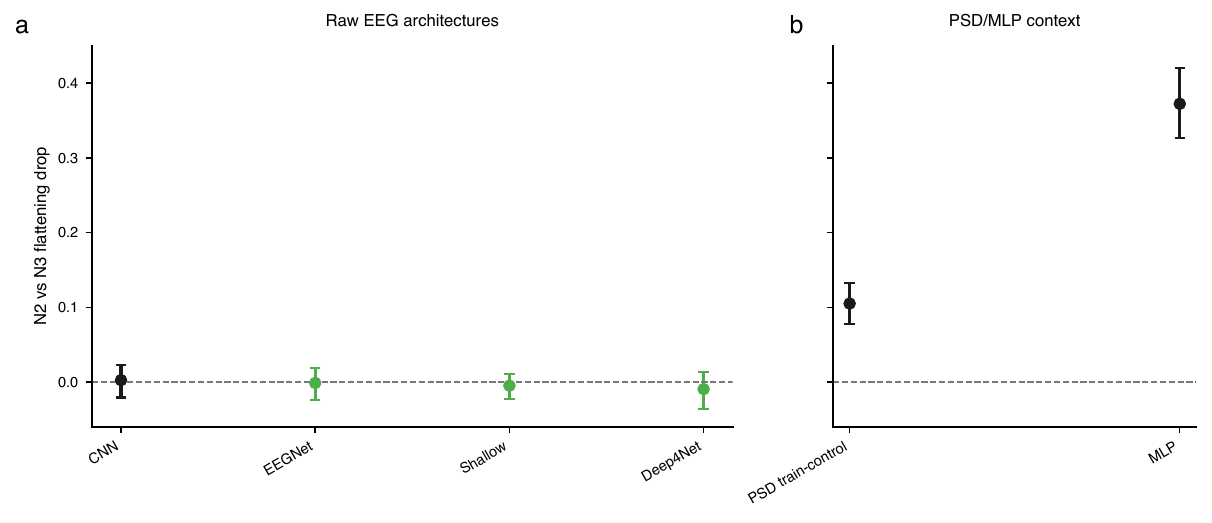"}
\caption{\textbf{Extended Data Fig.~2 $|$ Sleep N2-versus-N3 flattening pattern.} Architecture-wise flattening drops for N2-versus-N3, showing near-zero or negative drops and supporting the task-specific interpretation.}
\end{figure}
\clearpage

\begin{figure}[H]
\centering
\includegraphics[width=\textwidth]{"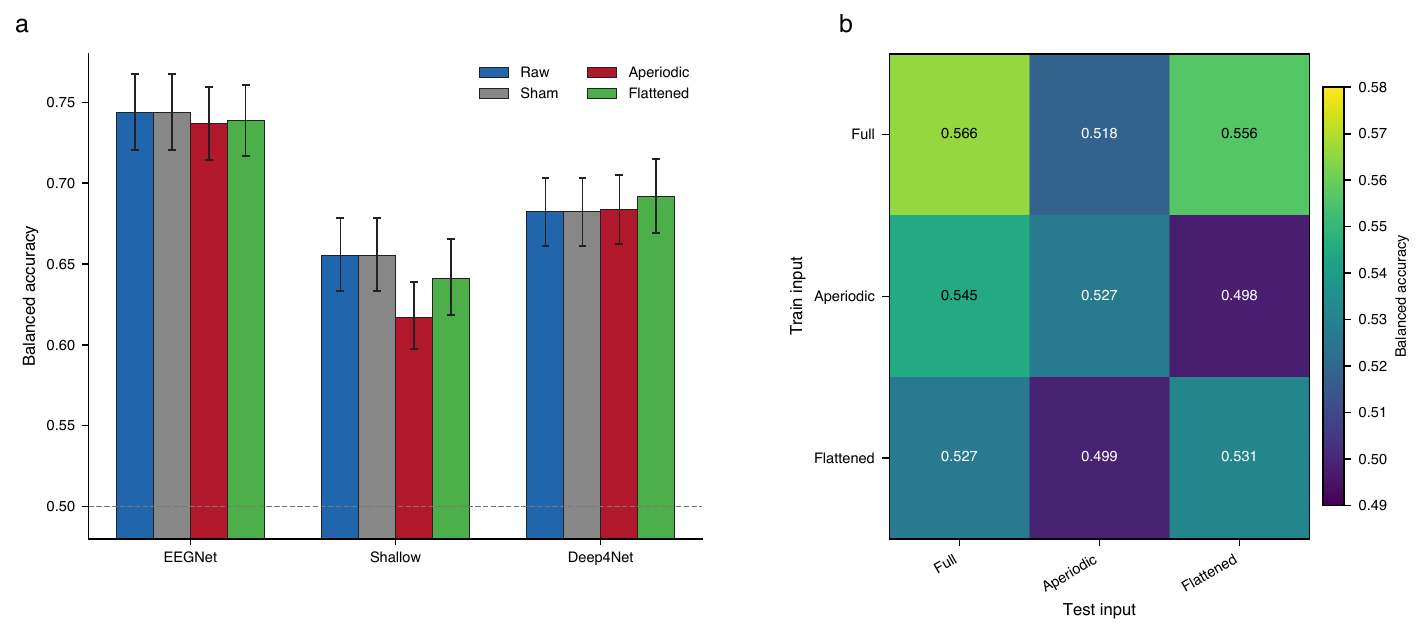"}
\caption{\textbf{Extended Data Fig.~3 $|$ PhysioNet MI full results.} Neural and PSD intervention results for PhysioNet motor imagery, including the train-by-test PSD matrix.}
\end{figure}
\clearpage

\begin{figure}[H]
\centering
\includegraphics[width=\textwidth]{"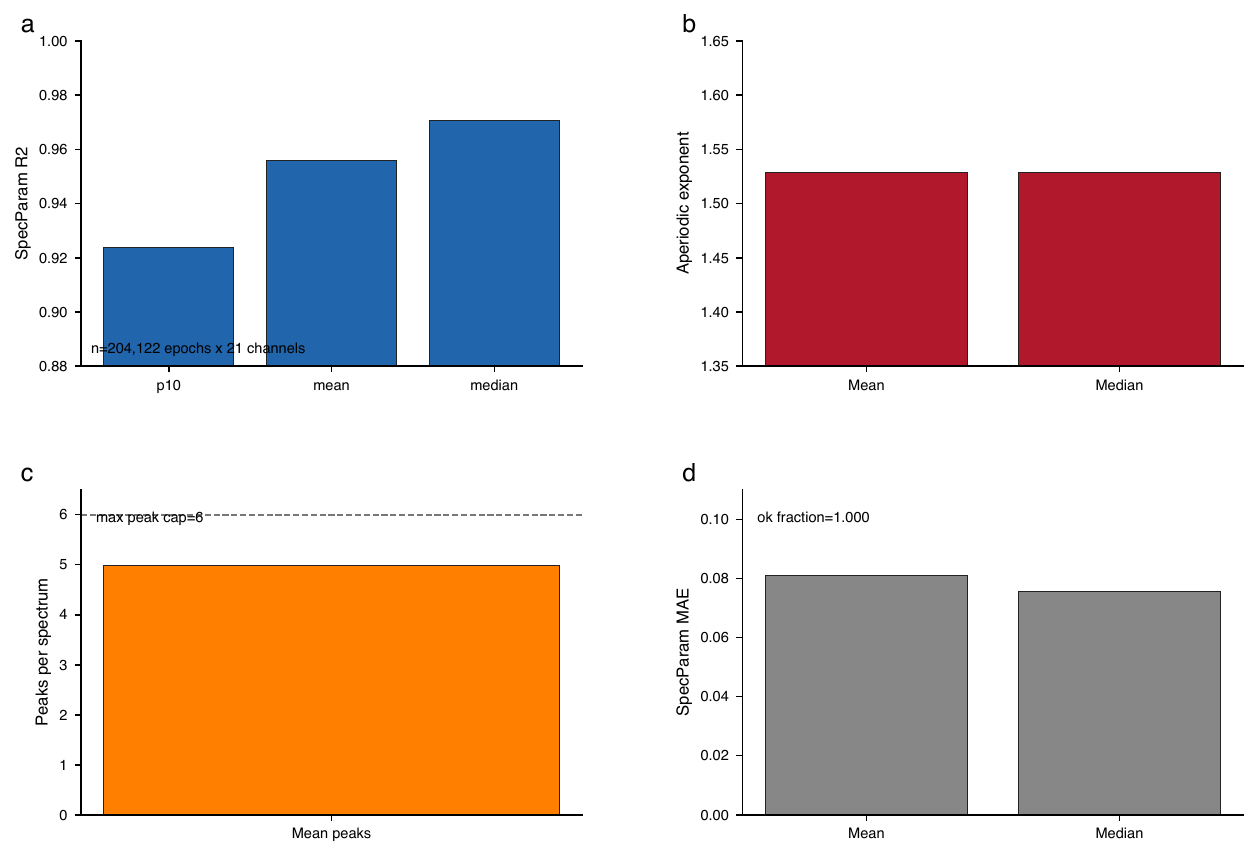"}
\caption{\textbf{Extended Data Fig.~4 $|$ Full-TUAB SpecParam fit quality.} Distributional summaries of full-corpus TUAB spectral fit quality, exponent estimates and peak-cap diagnostics.}
\end{figure}
\clearpage

\begin{figure}[H]
\centering
\includegraphics[width=\textwidth]{"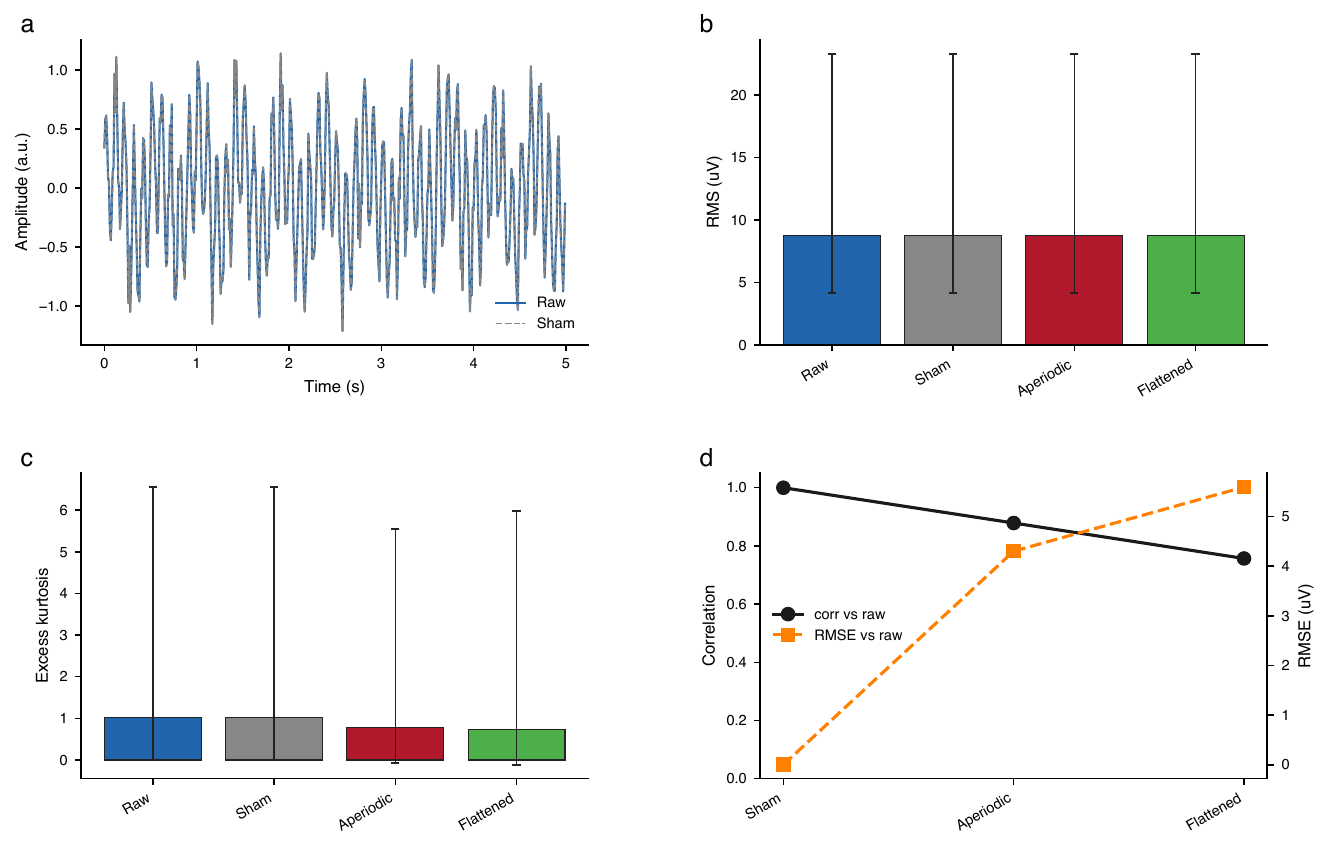"}
\caption{\textbf{Extended Data Fig.~5 $|$ Raw intervention diagnostics.} Time-domain and distributional diagnostics for raw EEG Fourier interventions, including sham reconstruction and amplitude-distribution summaries.}
\end{figure}
\clearpage

\begin{figure}[H]
\centering
\includegraphics[width=\textwidth]{"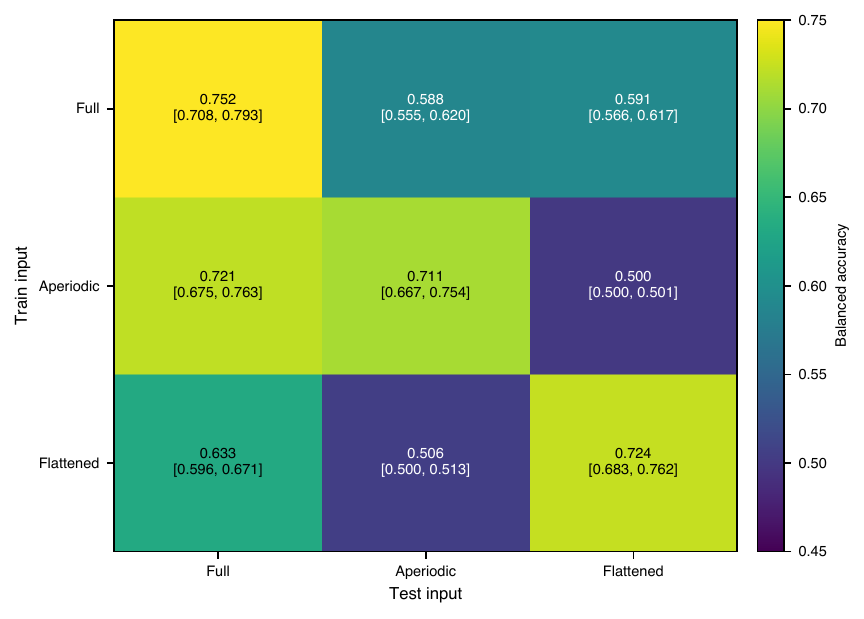"}
\caption{\textbf{Extended Data Fig.~6 $|$ Full-TUAB PSD intervention matrix.} Full-corpus TUAB PSD ridge train-by-test intervention grid.}
\end{figure}
\clearpage

\begin{figure}[H]
\centering
\includegraphics[width=\textwidth]{"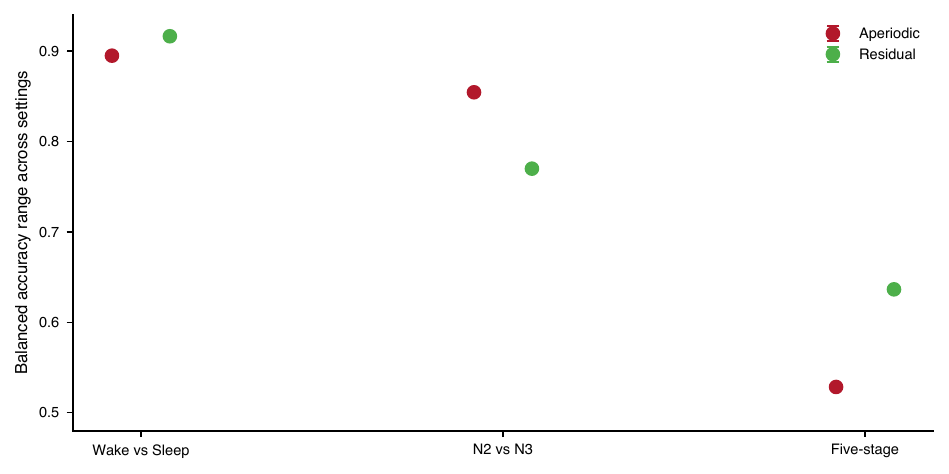"}
\caption{\textbf{Extended Data Fig.~7 $|$ Sleep-EDF SpecParam sensitivity.} Sensitivity of Sleep-EDF conclusions to decomposition settings.}
\end{figure}
\clearpage

\begin{figure}[H]
\centering
\includegraphics[width=\textwidth]{"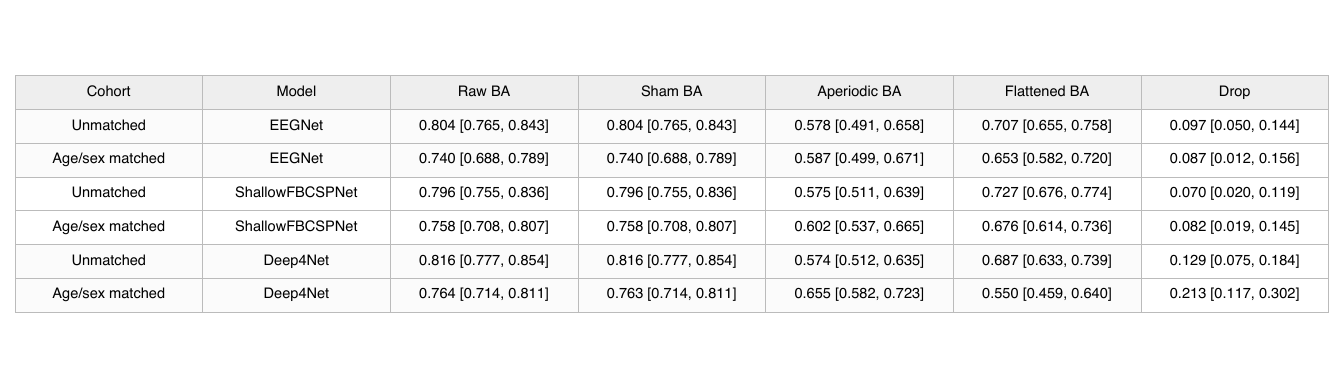"}
\caption{\textbf{Extended Data Fig.~8 $|$ Full-TUAB age-matched neural results.} Complete unmatched versus age/sex-matched full-TUAB neural intervention table across architectures; the matched control used 921 pairs, including 87 evaluation pairs.}
\end{figure}
\clearpage

\begin{figure}[H]
\centering
\includegraphics[width=\textwidth]{"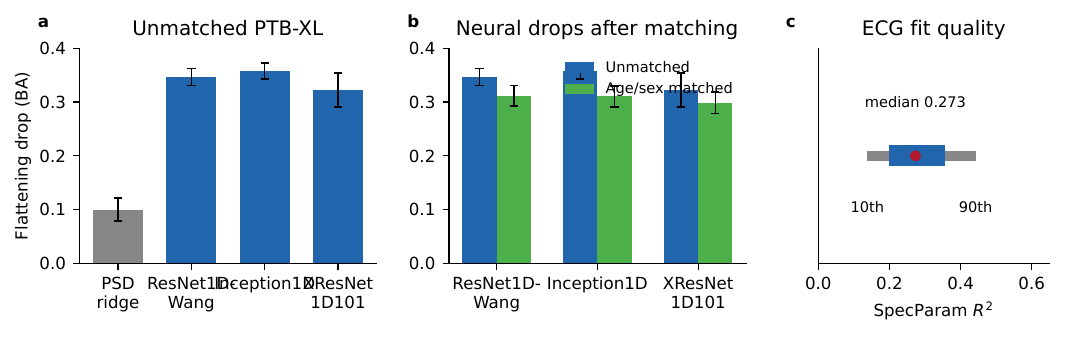"}
\caption{\textbf{Extended Data Fig.~9 $|$ PTB-XL ECG spectral audit results.} PSD ridge and neural architecture intervention results for PTB-XL normal-versus-abnormal ECG classification, including unmatched and age/sex-matched controls.}
\end{figure}
\clearpage

\begin{figure}[H]
\centering
\includegraphics[width=\textwidth]{"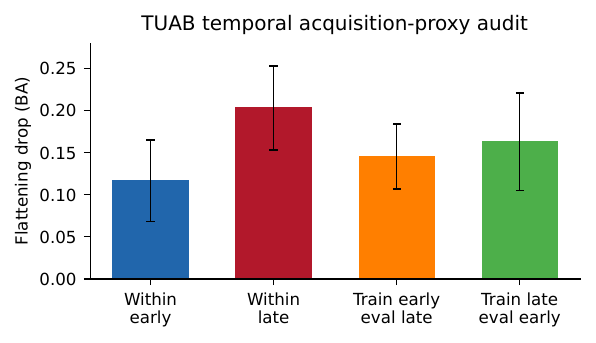"}
\caption{\textbf{Extended Data Fig.~10 $|$ TUAB temporal acquisition-proxy audit.} PSD ridge flattening drops within and across temporal recording bins (2009--2011 versus 2012--2013), showing that the aperiodic effect persists independently of recording era.}
\end{figure}

\clearpage
\section*{Supplementary Information}

The Supplementary Information contains complete numerical results, full-TUAB SpecParam fit-quality summaries, IRASA downstream ridge interventions, TUAB age metadata and matching details, the PhysioNet MI PSD intervention matrix, PTB-XL ECG audit tables, TUAB temporal acquisition-proxy audit tables, formal hypothesis tests, mathematical notes, simulation validation details and a proposed aperiodic reporting checklist.

\textbf{Supplementary Table 8 | Formal hypothesis tests with BH-FDR correction.} One-sided bootstrap p-values for flattening-drop tests and two-sided p-values for sham-control tests, with Benjamini--Hochberg FDR-corrected p-values at $q=0.05$, across 31 primary flattening comparisons and 28 sham comparisons.

\textbf{Supplementary Table 9 | PTB-XL ECG spectral audit numerical results.} Balanced accuracy and 95\% bootstrap confidence intervals for PSD ridge and neural PTB-XL audits, including unmatched and age/sex-matched controls.

\textbf{Supplementary Table 10 | TUAB temporal acquisition-proxy PSD audit.} PSD ridge flattening drops and bootstrap confidence intervals for within-era and cross-era TUAB recording-date analyses.

\section*{Acknowledgements}

Sleep-EDF, PhysioNet Motor Movement/Imagery and PTB-XL data are available through PhysioNet. TUAB access was provided by the Temple University Hospital EEG Corpus under their data-use agreement.

\section*{Author contributions}

J.S.B.\ conceived the study, designed and implemented the spectral audit framework, performed all experiments and analyses, and wrote the manuscript. S.P.\ supervised the research and reviewed the manuscript.

\section*{Competing interests}

The authors declare no competing interests.

\section*{Data availability}

Sleep-EDF, PhysioNet Motor Movement/Imagery and PTB-XL are publicly available through PhysioNet. TUAB is available from the Temple University Hospital EEG corpus, subject to data-use approval. Derived result tables used for the figures and Supplementary Information are stored in the project repository.

\section*{Code availability}

Code for preprocessing, model training, spectral interventions, and statistical aggregation will be made publicly available upon publication of the peer-reviewed version of this work.

% The sn-nature document-class option applies \bibliographystyle{sn-nature}.
\bibliography{references}

\end{document}